\documentclass[final]{cvpr}

\usepackage{times}
\usepackage{epsfig}
\usepackage{graphicx}
\usepackage{amsmath}
\usepackage{amssymb}

\usepackage{bbding}
\usepackage{booktabs}
\usepackage{hyperref}
\hypersetup{
    colorlinks=true,
    linkcolor=red,
    urlcolor=magenta,
}
\usepackage{bbm}

\def\cvprPaperID{****} 
\def\confYear{CVPR 2021}

\begin{document}

\title{Generic Event Boundary Detection Challenge at CVPR 2021 \\
Technical Report: Cascaded Temporal Attention Network (CASTANET)
}

\author{Dexiang Hong$^{1,\ast}$, Congcong Li$^{1,}$\thanks{Both authors contributed equally to this work.}, Longyin Wen$^2$, Xinyao Wang$^2$, Libo Zhang$^3$ \\
$^1$University of Chinese Academy of Sciences, Beijing, China.\\
$^2$Bytedance Inc. Mountain View, USA.\\
$^3$Institute of Software Chinese Academy of Sciences, Beijing, China.\\
{\tt\small \{hongdexiang19, licongcong18\}@mails.ucas.ac.cn} \\
{\tt\small \{longyin.wen, xinyao.wang\}@bytedance.com} \\
{\tt\small libo@iscas.ac.cn}
}

\maketitle

\begin{abstract}
This report presents the approach used in the submission of Generic Event Boundary Detection (GEBD) Challenge at CVPR21. In this work, we design a Cascaded Temporal Attention Network (CASTANET) for GEBD, which is formed by three parts, the backbone network, the temporal attention module, and the classification module. Specifically, the Channel-Separated Convolutional Network (CSN) is used as the backbone network to extract features, and the temporal attention module is designed to enforce the network to focus on the discriminative features. After that, the cascaded architecture is used in the classification module to generate more accurate boundaries. In addition, the ensemble strategy is used to further improve the performance of the proposed method. The proposed method achieves $83.30\%$ F1 score on Kinetics-GEBD test set, which improves $20.5\%$ F1 score compared to the baseline method. Code is available at \href{https://github.com/DexiangHong/Cascade-PC}{https://github.com/DexiangHong/Cascade-PC}.
\end{abstract}

\begin{figure*}
\begin{center}
\includegraphics[width=0.99\linewidth]{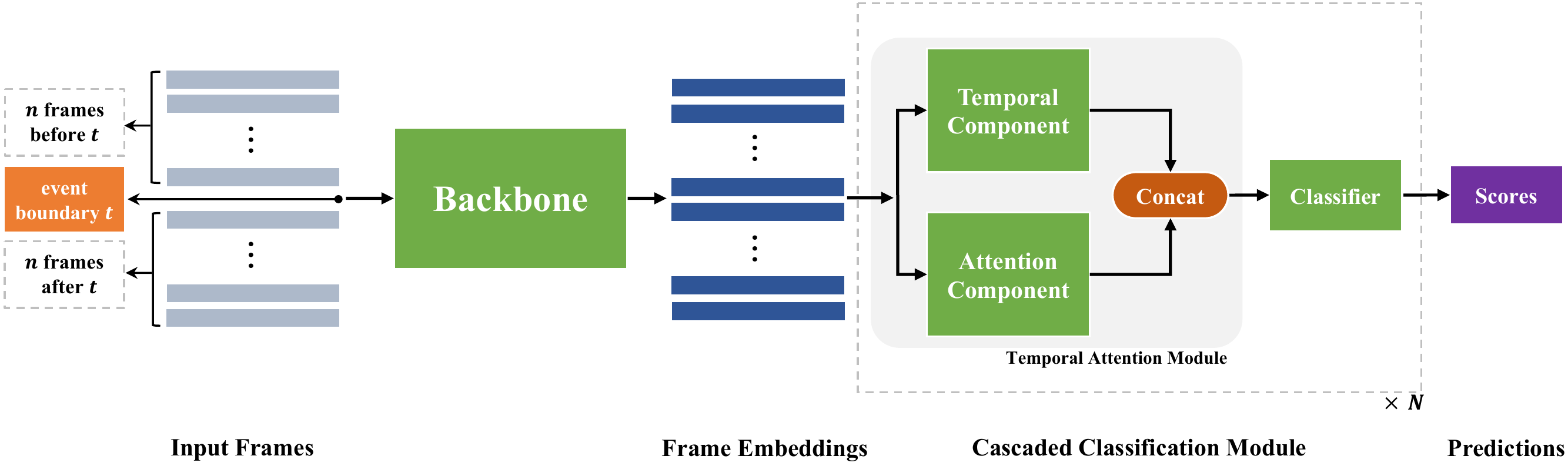}
\end{center}
  \caption{The architecture of the proposed cascaded temporal attention network.}
\label{fig:framework}
\end{figure*}

\section{Approach}
The task of \textbf{Generic Event Boundary Detection (GEBD)} is introduced by \cite{DBLP:journals/corr/GEBD_dataset}, which aims to localize the moments where humans naturally perceive taxonomy-free event boundaries that break a longer event into shorter temporal segments. The main challenge of GEBD is the taxonomy-free nature of event boundary, which requires the predictions of the detector matching at least one annotator. To tackle this problem, we design a Cascaded Temporal Attention Network (CASTANET), which consists of three parts, the backbone network, the temporal attention module and the cascaded classification module. The Channel-Separated Convolutional Network (CSN) \cite{CSN} is used as the backbone network to extract features. After that, the temporal attention module is designed to enforce the network to focus on the discriminative features. Finally, the cascaded classification module is used to generate the accurate boundaries. Notably, the ensemble strategy is used to further improve the performance of the proposed method. The overall architecture of the proposed method is presented in Figure \ref{fig:framework}. We will describe each module in more detail in the following sections.  

\subsection{Backbone Network}
The Channel-Separated Convolutional Networks (CSN) \cite{CSN} is first designed for video classification, which separates the channel interactions and spatiotemporal interactions to balance the accuracy and efficiency. Specifically, all the convolutional operations in CSN are separated into the pointwise $1\times1\times1$ convolutions for channel interactions and depthwise $3\times3\times3$ convolutions for local spatiotemporal interactions. To extract more discriminative features for GEBD, we use CSN as the backbone network to extract features.

\subsection{Temporal Attention Module}
\noindent\textbf{Temporal component.} Temporal context information is crucial to determine the event boundaries. Some event boundaries may need more contextual frames while others may need only a few frames to determine whether are boundaries. To increase the receptive field of different layers and capture contextual information of different resolutions, we stack several dilated 1D convolution layers in temporal domain. Inspired by the MS-TCN++ \cite{DBLP:journals/corr/MS-TCN++} architecture, we use a dilation factor that is doubled at each layer, \textit{i.e.} $\lbrace1,2,\cdots, 2^{L-1}\rbrace$, where $L$ is the total number of layers($L=4$ in our experiments). Each layer has the same number of convolutional filters and applies a dilated convolution with ReLU activation to the output of the previous layer. We further use residual connections to facilitate gradients flow. The set of operations at each layer can be formally described as follows:

\begin{equation}
	 \hat{C_l} = \text{ReLU}(W_d * C_{l-1}) ,
\end{equation}
\begin{equation}
	C_l = C_{l-1} + W * \hat{C_l} ,
\end{equation}
where $C_l$ is the output of layer $l$, $*$ denotes the convolution operator, $W_d \in \mathbb{R}^{3 \times D \times D}$ are the weights of the dilated convolution filters with kernel size 3 and D is the number of convolutional filters, $W \in \mathbb{R}^{1 \times D \times D}$ are the weights of a $1\times1$ convolution, and bias parameters are omitted for simplicity.

Meanwhile, we use a bi-directional LSTM \cite{DBLP:journals/neco/LSTM} to capture temporal information, which can process sequences of video frames in the temporal space. Then a 3d max pooling is applied on the LSTM output, which extracts the max activations among video frames at different time point. This operation flattens the sequential input video frames into the features $v_{tem} \in \mathbb{R}^c$ and maximize event boundary activations when there exists a boundary in the input video frames.

\noindent\textbf{Attention component.} GEBD requires to fully understand the semantics of each input video frame. Thus, it is necessary to learn the textural attention for boundary detection simultaneously. The Transformer \cite{DBLP:conf/nips/attention} is a model that uses self attention to boost the performance of neural machine translation. We adapt the same architecture as in \cite{DBLP:conf/nips/attention} and stack 4 self attention layers to process the input frames. Then a 3d max pooling is also applied on the attention output, producing the attention features $v_{att} \in \mathbb{R}^c$.

\begin{figure}
\begin{center}
\includegraphics[width=0.9\linewidth]{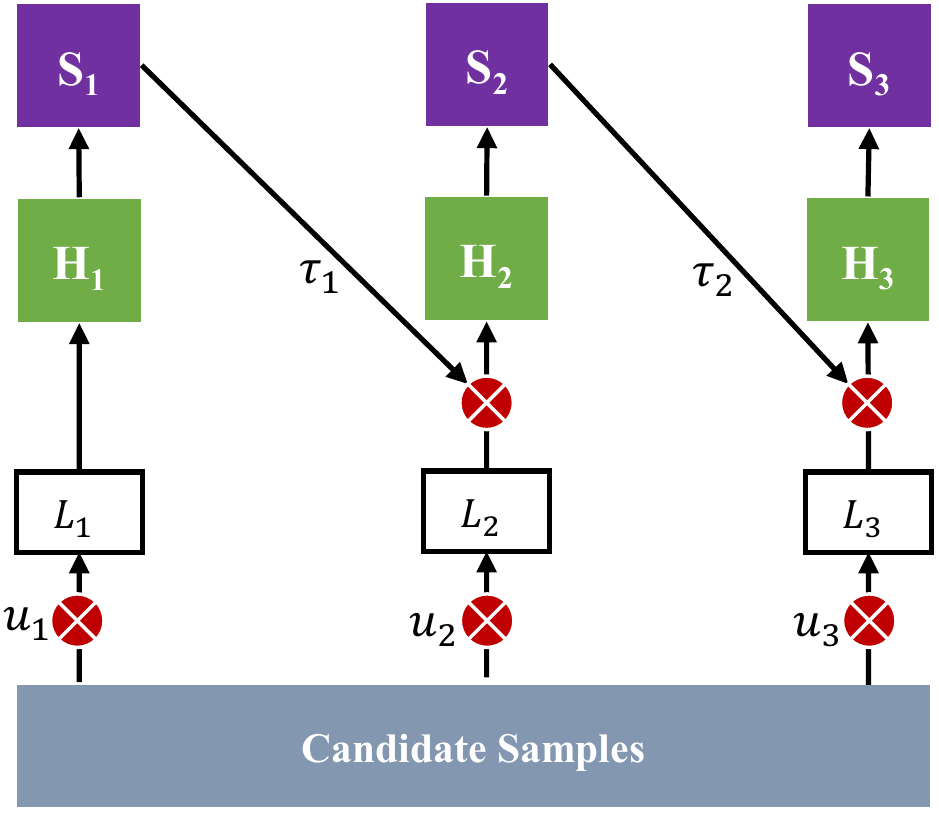}
\end{center}
   \caption{The architectures of cascaded classification module.``H" indicates classification head, ``L" indicates ground truth labels and ``S" indicates the classification scores. $u$ and $\tau$ are the ground truth threshold and mask threshold, respectively.}
\label{fig:cascade_head}
\end{figure}

\subsection{Cascaded Classification Module}\label{sec:cascaded_classifier}
Taking the extract features from the temporal attention module, we use a fully connected layer to produce the classification scores of the boundaries, which is computed as follows:
\begin{equation}
    S = \text{softmax}(\text{FC}([v_{tem}, v_{att}])),
\end{equation}
where $v_{tem}$ and $v_{att}$ are the features produced by the temporal and attention components, and $[\cdot, \cdot]$ indicates concatenation in the channel dimensions.

Improving both recall and precision for GEBD is a dilemma. Since the event boundary is taxonomy-free, it is difficult to determine if a frame is the event boundary or not. The official evaluation protocol \cite{DBLP:journals/corr/GEBD_dataset} addresses this problem by using \textit{Rel.Dis.}, which indicates the error between the detected and ground truth timestamps, divided by the length of the corresponding action. Then given a fixed threshold $u$ for \textit{Rel.Dis.}, we can determine whether a detection is correct (\textit{i.e.} $\leq$ threshold) or not (\textit{i.e.} $>$ threshold). When $u$ is high, the positive samples contain more ambiguous frames, and make the classifiers confused and have little incentive to reject close false positives. When $u$ is low, the positive is more accurate, but it is difficult to assemble enough positive training examples. Inspire by Cascade R-CNN \cite{DBLP:conf/cvpr/cascade-rcnn}, we propose the cascaded architecture to gradually refine the classification results. The cascaded classifies are trained sequentially, using the output from the previous stages.

Specifically, given a sequential of classification heads $\lbrace H_1,H_2,\cdots,H_N\rbrace$ of identical architecture (gray dashed box in Figure \ref{fig:framework}, initialized with different parameters) where $N$ is the number of cascaded heads, we train these classifier heads using different ground truth labels $\lbrace L_1,L_2,\cdots,L_N\rbrace$ which are produced by decreased ground truth \textit{Rel.Dis.} threshold $\lbrace u_1,u_2,\cdots,u_N\rbrace$, where $u_1 \geq u_2 \geq \cdots \geq u_N$. Then the output classification score $S_n$ of $H_n$ is used as a mask with threshold $\tau_{n}$ to filter out easy negative samples and the next stage detection head $H_{n+1}$ further refines positive samples using previous head's mask by utilizing more accurate ground truth labels $L_{n+1}$. The mask thresholds $\lbrace \tau_1,\tau_2,\cdots,\tau_{N-1}\rbrace$ used to filter out easy negative samples are monotonically increasing, \textit{i.e.}, $\tau_1 < \tau_2 < \cdots < \tau_{N-1}$. The cascaded classifier architecture is shown in figure \ref{fig:cascade_head}.

During inference, the final output is computed by averaging all the outputs of the detection heads.

\subsection{Audio Features}
Joint audiovisual learning is the core to human perception. For many video understanding tasks, audio could be very helpful \cite{avslowfast}. We extract the audio from the original videos and use STFT to convert the audio representation into the frequency domain. Then four 1D convolution layers are adopted to extract the temporal feature of the audio inputs. Then we concatenate the audio feature with visual features and feed them to the linear classification layers.

\section{Experiments}

\subsection{Implementation Details}
\textbf{Dataset.} We randomly sample $2000$ videos from kinetics-GEBD validation set to construct local validation set and use all the other data for training. 

\textbf{Network architecture.} We use Channel-Separated Convolutional Networks (CSN) \cite{CSN} pretrained on the IG-65M \cite{IG} and kinetics-400 as the feature encoder. The pretrained weight is released in the mmaction2 repository\footnote{\url{https://github.com/open-mmlab/mmaction2}}. Specifically, we modify the temporal strides of the CSN backbone from $[1, 2, 2, 2]$ to $[1, 1, 1, 1]$. As a result, The CSN backbone produces feature embeddings with the original temporal resolution. We use 4 dilated temporal convolutional layers with dilation rates $[1, 2, 4, 8]$ followed by 1 bi-directional LSTM layer in temporal module and 4 self attention layers \cite{DBLP:conf/nips/attention} in attention module. The outputs of temporal module and attention module are concatenated and then fed into a fully connected layer to make prediction. For each training sample, the input dimension is $2n\times 224 \times 224$, where $n=8$ is the number of frames before and after the candidate time position $t$, $224$ is the height and width of the resized frame.

\begin{table}[!t]
    \centering
    \begin{tabular}{c c c c}
    \toprule[1.2pt]
        Model & F1 & Precision & Recall \\
    \hline
        Baseline\cite{DBLP:journals/corr/GEBD_dataset} & 0.615 & 0.584 & 0.684 \\
        + CSN \cite{CSN} & 0.719 & 0.645 & 0.811 \\
        + Dynamic Sampling & 0.750 & 0.682 & 0.831 \\
        + Temporal Attention & 0.778 & 0.719 & 0.847 \\
        + Cascade Classifier & 0.782 & 0.739 & 0.832 \\
        + Audio & 0.789 & 0.734 & 0.853 \\
        + Ensemble & 0.814 & 0.776 & 0.868 \\
    \bottomrule[1.2pt]
    \end{tabular}
    \caption{The results of ablation studies on our local validation set. Here the baseline is the pairwise boundary classifier (PC) proposed in \cite{DBLP:journals/corr/GEBD_dataset}}
    \label{tab:ablation}
\end{table}

\begin{table}[!t]
    \centering
    \begin{tabular}{c c c c}
    \toprule[1.2pt]
        Model & F1 & Precision & Recall \\
    \hline
        Baseline\cite{DBLP:journals/corr/GEBD_dataset} & 0.625 & 0.624 & 0.626 \\
        \textbf{Ours} & 0.833 & 0.838 & 0.828 \\
    \bottomrule[1.2pt]
    \end{tabular}
    \caption{The final results on the test set}
    \label{tab:final}
\end{table}

\textbf{Dynamic sampling.} Videos in the Kinetics-GEBD dataset are recorded in different FPS. Different from the baseline method that samples $1$ frame for every $3$ frames, we determine the sampling frequencies based on the recorded FPS of videos. Notably, given the video recorded with $f$ FPS, we sample $1$ frame for every $\lceil{f/8}\rceil$ frames. Intuitively, the larger the FPS is, the more repeated frames it contains. So fixed frame sampling strategy may include redundant information for larger FPS while lack necessary contextual information for smaller FPS. Our dynamic sampling strategy captures contextual information with a fixed ratio, which makes input information consistent and is benefit for network convergence.

\textbf{Training.} We train our model on 8 Tesla V100 GPUs using PyTorch 1.8 and use a mini-batch of 32 clips per GPU, thus making a total mini-batch of 256 clips. Training is done in 1 epoch since the performance drops if training longer.  We use SGD optimizer with the momentum of $0.9$ and the weight decay of $5e^{-4}$. The learning rate is set to $1e^{-2}$. Automatic mixed precision training\footnote{\url{https://pytorch.org/docs/stable/notes/amp_examples.html}} is also adapted to reduce GPU memory burden.

\textbf{Testing.} For each candidate boundary time $t$, we average all the scores of the cascade classifiers which are trained using different threshold-level ground truth labels presented in section\ref{sec:cascaded_classifier}. Then we watershed the probability sequence to obtain internals above 0.5 and each internal’s center is treated as an event boundary.

\subsection{Ablation Studies}
In this section, we conduct several ablation experiments to comprehensively understand the performance contribution of the proposed framework. The contributions of each module are shown in Table \ref{tab:ablation}. We report the results on the local validation set. We can observe that with these methods, we improve the F1 score from 0.615 to 0.814. Specifically, we set cascade classifier's mask threshold $\tau$ to [0.4, 0.3] and ground truth \textit{Rel.Dis.} threshold $u$ to [0.5, 0.4, 0.3]. Finally, we finetune the mask threshold and ground truth \textit{Rel.Dis.} threshold and ensemble 9 models to produce the final results.

\subsection{Overall Results}
We submit our results on the Kinetics-GEBD test set and achieve $83.30$ F1 scores. Specifically, we ensemble $9$ models with different parameters. The results are shown in Table \ref{tab:final}.



{\small
\bibliographystyle{ieee_fullname}

}

\end{document}